**Karolina Rudnicka**[1]
**University of Gdańsk**

**The "negative end" of change in grammar:**
**Terminology, concepts and causes**

**Abstract**

The topic of "negative end" of change is, contrary to the fields of innovation and emergence, largely under-researched. Yet, it has lately started to gain an increasing attention from language scholars worldwide. The main focus of this article is threefold, namely to discuss the i) terminology; ii) concepts and iii) causes associated with the "negative end" of change in grammar. The article starts with an overview of research conducted on the topic. It then moves to situating phenomena referred to as *loss*, *decline* or *obsolescence* among processes of language change, before elaborating on the terminology and concepts behind it. The last part looks at possible causes for constructions to display a (gradual or rapid, but very consistent) decrease in the frequency of use over time, which continues until the construction disappears or there are only residual or fossilised forms left.

1. **Introduction**

This is a theoretical and conceptual paper on the "negative end" of change in language, which, having an opinion of being under-researched (e.g. Fritz 2006; Hundt 2014: 166; Tichý 2018:

---

[1] https://orcid.org/0000-0001-8097-6086
University of Gdańsk, Faculty of Languages
karolina.rudnicka@ug.edu.pl

81), has recently been gaining an increasing attention from the research community (e.g. Rudnicka 2018, 2019; Kranich and Breban forthcoming; Elsweiler and Huber forthcoming).

The process in which constructions and lexical items move towards this end of change and the very final stage of their existence, are, in the research literature, referred to in various ways, for instance (in alphabetical order) *breakdown*, *collapse*, *death*, *decline*, *demise*, *disappearance*, *end*, *loss*, *obsolescence*, *obsoleteness*, *rareness*. Sometimes authors use these notions interchangeably, as synonyms. One of the aims of the this paper is to discuss both the terminology and the theoretical concepts behind it, hopefully making this complex picture a bit more ordered. This topic is, however, carried over to Section 4, after a survey of literature is presented. Until then, the notions used in this work to refer to the "negative end" of change will include *loss* (for already cases already *lost*), and *decline* and *obsolescence* (for *loss* in progress).

Another goal of this work is to present a comprehensive but concise survey of research that has and is being done on the topic in question, with information on i) the studied variables and languages; ii) methods applied by the scholars; iii) theoretical assumptions formulated by different scholars. This survey is being split into two parts – Section 2, which looks at particular case studies and Section 3, dealing with theoretical assumptions pertaining to the topic of loss and described in the literature. Even though this paper focuses on the English language, it also mentions research done on other languages as it is assumed that the phenomena and processes described are likely to be generalizable across different languages.

Section 4, which has already been mentioned, presents various possibilities and proposes a few criteria for differentiation between constructions which are e.g. drastically declining in the frequency of use, and the ones, which are already lost (based on Rudnicka 2019).

The third goal, partly resulting from the two previous aims, and drawing from the research literature, is a discussion of potential causes for loss or decline in the realm of grammar, presented in Section 5. This boils down to the question on whether it is possible to point to concrete causes and reasons to account for a situation in which grammatical items leave the language system.

## 2. Concise survey of research literature

For a very long time the "negative end" of change in language remained under-researched with very few studies focusing on the decline or loss of concrete structures or lexical items, such as:
- Ashby (1981) – the decline of the negative morpheme *ne* in Parisian French;

"Negative end" of change in grammar● Hiltunen (1983) – the loss of prefixes and the emergence of the phrasal verb in Old and Early Middle English texts;

● Petré (2010) – the end of weorðan in the past tense in Old and Middle English;

● Hundt and Leech (2012) – the demise of the conjunction *for* and the relativiser *which* in British and American English;

● Hilpert (2012) – the decline of the *many-a*-noun construction in American English;

● Hundt (2014) – the decline of the *being to VERB* construction in British and American English;

● Kastronic & Poplack (2014) – the (only) apparent revival of mandative subjunctive in spoken (North) American English.

All of these studies apply methods of corpus linguistics, apart from data and frequency analysis also e.g. collexeme analysis (Petré 2010) and collostructional analysis with variability-based neighbour clustering (Hilpert 2012). Apart from these works, the topic of loss is mentioned, often in an anecdotal way (Mair 2006: 141), by e.g. Barber (1964: 130-131), Kortmann (1996), Lorenz (2013: 125-126).

Interestingly, the topics of loss and decline seem to attract more attention in the very recent time, compared to how they did in the past. One of the manifestations of this trend is the recent book by the Rudnicka (2019) devoted solely to the topic of obsolescence of grammatical structures. Another one is the collective volume edited by Kranich & Breban (forthcoming) following the workshop "Lost in change: causes and processes in the loss of grammatical constructions and categories" which took place in Stuttgart in 2018. The volume will contain papers focusing on different kinds and aspects of loss, both grammatical and lexical, among others by: Kempf, on the loss of the relative particle *so* in the early stages of (written) New High German; Kuo, analysing the loss of the adverse avertive schema in Mandarin Chinese; Rudnicka, on the English *so*-adj-*a* construction seen as a case of obsolescence in progress (forthcoming 01).

The rise of interest in the topic is further confirmed by recent works by:

○ Blanco-García (2017), dealing with the ephemeral concessive subordinators in Late Middle English and Early Modern English;

○ Rudnicka (2018), relating the decrease of overall sentence length in the nineteenth and twentieth century in (written) English to the obsolescence of the purpose subordinator *in order that*;

○ Tichy (2018), investigating lexical obsolescence in Late Modern and Present Day English;

○ Imel (2019), examining leftward stylistic displacement in Medieval French;

○ Sommerer and Hofmann (2020), looking at the functions and development of the determinative *sum(e)* in Old, Middle and Early Modern English;

○ Elsweiler and Huber (forthcoming), studying the loss of number in the English second person pronoun, which came about by the eighteenth century.

### 3. Processes of language change and the "negative end"

When it comes to research on the theory behind language change, disappearance of grammatical structures has been looked at from many different angles by e.g. Hopper and Traugott (2003); Haspelmath (2004); Lass (1990); Greenberg (1991); and Hansen (2017). Below, I present the main points raised by different authors.

● Hopper and Traugott (2003: 172) associate loss with competition. Furthermore, they notice that the declining forms remain in formal, written registers, while being (almost) completely absent from the colloquial language (Hopper and Traugott 2003: 173; Hundt 2014). An example of decline, which they give is the German *Imperfekt*. Another point Hopper and Traugott make, is that a dying form is often replaced by a newer (typically periphrastic) one. This scenario is referred to as "renewal".

● Lass (1990) introduces the notion of *exaptation* to account for a situation in which an item leaves the system of the core grammar and is recycled to serve a new purpose. In Lass's terminology (1990: 100) the very rare grammatical items with limited productivity, used only occasionally, are referred to as *nonaptations*. With regard to a very similar scenario, Greenberg (1991) suggests the name *regrammaticalization*. Interestingly, Lass divides the languages into „wasteful" and "conservationist" ones, depending on what happens with the no-longer-needed pieces of grammatical matter.

● Haspelmath (2004: 33) claims that in a prototypical case of *expansion* in grammaticalization (see Fig. 1a), loss of the older forms is something natural. With regard to scenarios in which the forms disappearing are not the oldest ones, the notions of *retraction* (see Fig. 1b) and *antigrammaticalization* (see Fig. 1c) are introduced. The difference between the former and the latter lays in the fact that in antigrammaticalization cases the grammaticalization chain, also referred to as a grammaticalization cline and defined as a series of transitions between overlapping stages of grammaticalization, expands to the left, contrary to what happens during the process of retraction. Interestingly, Haspelmath also claims that "[e]verything in language can become obsolete, independently of its degree of grammaticalization" (2004: 33).

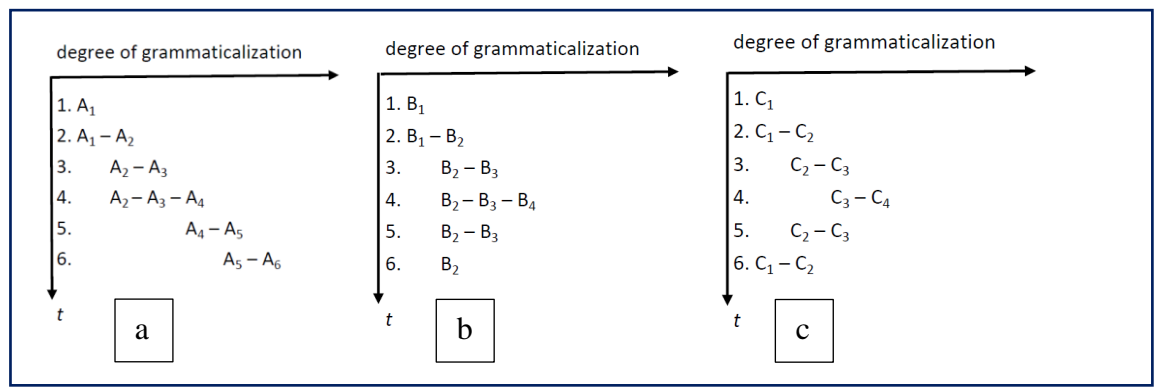

Fig. 1: Expansion (a), retraction (b) and antigrammaticalisation (c) (adapted from Haspelmath 2004: 33-34); t-axis refers to time.

● Hansen's (2017) work, which focuses on case studies from four Slavonic languages (Russian, Polish, Czech, and Serbian/Croatian) proposes "a typology that includes at least six processes: secondary grammaticalization, marginalization, degrammaticalization, retraction, lexicalisation, and grammatical word derivation". *Marginalization*, which, as a phenomenon, used to receive little attention from scholars (Hansen 2017: 278), is understood as process which "does not lead to the rise of an unmarked, highly frequent grammatical operator, but to elements which occupy a peripheral position in the language system, i.e. which are either stylistically restricted or co-occur with a limited number of verbs" (2017: 264). Thus, elements occupying the margins of the language system are not seen as forms resulting from grammaticalization but from the process of marginalization. This process might also be seen as a stage prior to retraction in the understanding of Haspelmath (2004: 34).

● Rudnicka (2019: 210) presents *grammatical obsolescence* as a process in its own right. In its course previously popular and productive constructions are, "often gradually, losing their productivity and popularity over time until they disappear or there are only residues or fossilised forms left". The function of an obsolescent construction might either continue to be (at least partially) expressed by an alternative construction, or discontinue. The work introduces a few criteria for identification of obsolescence and i.a. also links loss with overrepresentation of a construction in formal genres. The most striking finding of the study is, however, the fact that obsolescence is not attributed to constructional competition, but to so-called higher-order processes (2019: 212), understood as changes happening above the constructional level.

To sum up, the key findings standing behind the observations described in the present Section, can be summarised as follows:

● Expansion, marginalization, obsolescence, retraction and antigrammaticalization characterise the evolution within which the "negative end" changes take place;
● Exaptation, nonaptation, fosillization and regrammaticalization pertain to the future role of an obsolescent construction, given it will not disappear completely;
● Competition and higher-order processes are viewed as potential causes (of decline and loss);
● Concentration in particular (usually formal) genres or registers might be a symptom;
● Replacement is likely to happen, if the function is still needed.

## 4. Terminology and concepts

As mentioned in the Introduction, the *process* in the course of which grammatical constructions (or lexical items) leave the language system, is referred to in different ways. The same goes for the *stage* in which they already reach the end of their existence in language. Some works mention a *demise* (e.g. Hundt 2014), *decline* (e.g. Hiltunen 1983) or *obsolescence* (e.g. Rudnicka 2019; Kempf forthcoming) of a given construction, some investigate the *loss* (e.g. Elsweiler, Huber forthcoming) of certain items or features of the language. With regard to a complete *disappearance* or complete *loss* (e.g. Petré 2010), the notion of *death* (or a construction being *dead*) is sometimes used (e.g. Hundt 2014). Where a larger organizational unit is meant, one can find works describing a *breakdown* of e.g. a certain system (Petré 2010).

      Work by Rudnicka (2019) might be seen as an attempt to offer some terminological and conceptual ordering in this complicated picture. Table 1 presents an overview of reflections on the terminology. There are three main expressions selected to refer to three concepts represented by different situations associated with the "negative end", namely *obsolescence*, *obsoleteness* and *rareness*. For each of these notions, the last column of Table 1 proposes synonyms based on the research literature presented above. Furthermore, the Table lists three criteria which can be used to differentiate between these three concepts: i) the presence of negative correlation between time and the frequency of use attested in a given time period; ii) lack of acceptability by the language users; and iii) the presence of low frequency in the last decades of the investigated period. The iii) criterion is true for all of the situations, as it seems logical that no high-frequency construction should be perceived as one which is "on the way out". On the other hand, i) and ii) are essential to differentiating between an *obsolescent* or an *obsolete* construction. The former one should be showing a negative correlation between the frequency

of use and time (it is seen as a necessary condition[2]), which can be tested with the use of Kendall's tau test[3], at the same time staying fully acceptable by the language community. In contrast to that, the latter one is no longer functional and acceptable for the language users and it does not need to show a decrease in the frequency of use. Thus, the adjective *obsolescent* refers to a construction undergoing a process (of obsolescence), whereas being obsolete is a description of a state. The criterion of acceptability is based on Hundt (2014: 185), who suggests using acceptability ratings for grammaticality determination in the case of uncommon or very infrequent constructions. Hence, it is also used to differentiate between obsolete and *rare* (having a low frequency of use) constructions. Such low frequency might result from i) obsolescence or ii) be one of the construction's traits, as some grammatical items never reach very high frequency of use or popularity (e.g. possibly being a result of e.g. marginalization, see Section 3). Furthermore, the concept of a construction being rare – or having a low frequency – does contain a component of relativity and arbitrariness – there is no universal frequency threshold dividing the constructions or words into frequent and rare ones.

| Adjective and noun | Meaning | Negative correlation between time and the frequency of use | Lack of acceptability by the language users | Low or none frequency in the last decades of the investigated period | Possible synonyms: adjectives and nouns |
|---|---|---|---|---|---|
| *obsolescent* | In the process of leaving the core grammar of a language | yes – it is a necessary condition ✓ | no ✗ | yes ✓ | *declining* |
| *obsolescence* | | | | | *decline* |
| *obsolete* | A state of already being outside of | no, not necessarily | yes – it is a necessary condition | yes | *lost* *dead* *extinct* *forgotten* |

---

[2] As described by Shustack (1988: 93), "a necessary condition for some outcome is a condition that must always be present for the outcome to be present, that is a condition whose absence will prevent that outcome."
[3] Both Spearman's rho and Kendall's tau might be good for this purpose, here, Kendall's tau is chosen, due to its interpretational simplicity (Rudnicka 2019: 71).

| | | | | | |
|---|---|---|---|---|---|
| **obsoleteness** | the core grammar | ✗ | ✓ | ✓ | *death demise disappearance extinction loss* |
| **rare** | A state of being infrequent; having a low frequency of use | no, not necessarily ✗ | no ✗ | yes ✓ | *Infrequent marginal uncommon* |
| **rareness** | | | | | *infrequency* |

Table 1: Terminology pertaining to the "negative end" of change (based on Rudnicka 2019: 24).

Such threshold, could, at least theoretically, be helpful in classifying constructions as e.g. *rare* or *frequent* or *popular*, however, its determination does not seem possible, as Hundt (2014: 185) claims, "[o]ne of the questions that corpus evidence on its own cannot answer is what the frequency threshold of a construction is for it to be considered part of the grammar of individuals or of a speech community".

Fig. 2 shows symbolic representation of three different situations: obsolescence – a); rareness – b); and *fluctuations* – c). In the case of fluctuations, the construction is neither obsolescent, nor obsolete. The black, diamond-shaped points visible on the frequency curves in a) and b), symbolise random points in time, at which acceptability ratings can be conducted to check the acceptability of a given construction (to test whether the construction is already obsolete).

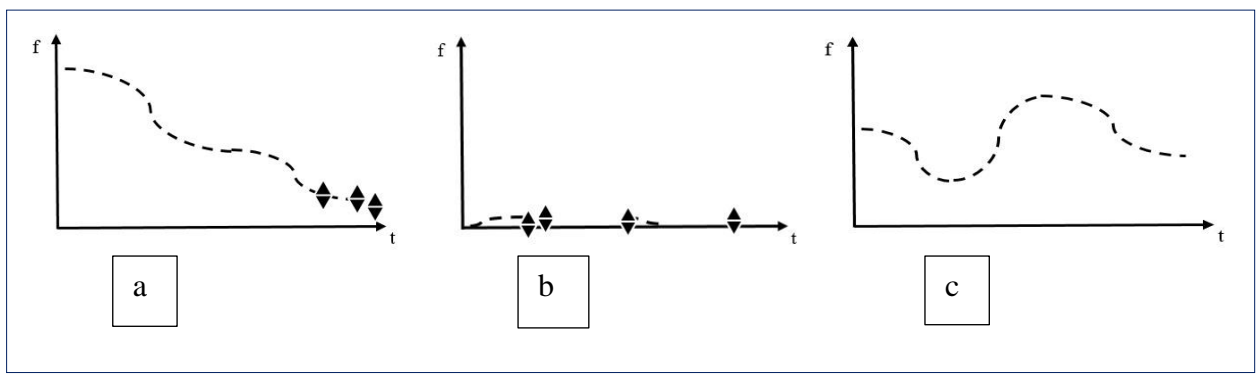

Fig. 2: Symbolic depictions of frequency (f-axis) across time (t-axis) in a potentially obsolescent construction (a); a rare construction (b); a construction the frequency of which shows fluctuations (c), based on Rudnicka (2019: 11).

Apart from the presence of negative correlation, two more possible symptoms of obsolescence or demise of constructions have been noticed in the literature. Among them, there is distributional fragmentation (termed so by Leech et al., 2009) referring to a situation in which the given construction is overrepresented in a given, mostly formal and written genre or a particular register (see also Section 3 of this paper). Importantly, distributional fragmentation is to be seen as a process – the form should be "increasingly restricted to certain genres" (Leech et al. 2009: 81). One example of distributional fragmentation at work is the purpose subordinator *in order that*, which shows a growing restriction to *non-fiction* genre (as represented by COHA[4] data), as modelled[5] by Rudnicka (2019: 137). Another symptom is paradigmatic atrophy, which, bearing some similarity to distributional fragmentation, pertains to morphological forms of a given construction. It can be described as a process in which lexical or grammatical items are increasingly restricted to particular morphological forms and certain syntactic environments (Hundt 2014: 170; Rudnicka 2019: 155). One example of paradigmatic atrophy is "the increasing rarity of the negative contraction in *–n't* with some (though not all) modals" (Leech et al. 2009).

Summing up, there are at least three different symptoms typical for constructions which are likely approaching "the negative end" of their constructional existence: a) the presence of negative correlation between time and the frequency of use; b) distributional fragmentation; and c) paradigmatic atrophy. Let us move on to the possible causes.

## 5. "Negative end" of change - causes

Obsolescence is, in the literature, associated with two kinds of phenomena

---

[4] The Corpus of Historical American English (COHA): 400 million words, 1810-2009. Available online at https://corpus.byu.edu/coha/.
[5] Extrapolated frequencies of occurrence are used to calculate the probability of coming across a given construction in each of the COHA genres (*fiction*, *newspaper*, *magazine* and *non-fiction*), see Rudnicka (2019: 136).

● competition (Hopper and Traugott; Hundt & Leech 2012), when we think about loss of form and its replacement. It is often referred to as an important factor influencing any language change (e.g. Lass 1997; Rissanen 2007; Barðdal & Gildea 2015) and is defined as a situation in which there are (at least) two forms competing for the same functional niche in the language. According to Lass (1997) the forms need to be functionally equivalent and the competition (and possible replacement) should be reflected in the frequency of use of the forms in question (Hundt and Leech 2012: 176).

● so-called system dependency (Hiltunen 1983; Petré 2010; Hundt 2014), also referred to as higher-order processes (Rudnicka 2019; Kempf forthcoming, Rudnicka forthcoming). It might cause obsolescence of 1) form only; or 2) form and function. Higher-order processes are defined as changes concerning a higher organizational level than the construction (Hilpert 2013: 14).

Rudnicka (2019; forthcoming), suggests that higher-order processes manifest themselves by frequency changes which, at the first sight, might resemble competition. An example is provided by some of the developments visible in the network of English purpose subordinators in Late Modern and Present Day English (Rudnicka 2019; forthcoming). Fig. 3 presents a frequency evolution of *in order that* (1) and its assumed competitor *in order for * to* (2). Still, a closer look at the data, history and other developments in the network (for a detailed analysis of the case of *in order that*, see Rudnicka forthcoming), suggests that there are at least two higher-order processes behind the drastic loss of frequency seen in Fig. 3, which are further categorised into *externally*- and *internally*- motivated ones[6].

An example of an externally-motivated higher-order process are the drastic socio-cultural changes of the nineteenth and twentieth century (Rudnicka 2019: 182), such as the advent of mass literacy, mass circulation newspapers and the development of new printing technologies. These high-impact socio-cultural changes led to informalization of the language of the media (Leech et al. 2009) and colloquialization (Mair 1998), which, in turn, are shown to have influenced the supraconstructional layer of the language causing e.g. a decrease in sentence length (Rudnicka 2018) and e.g. dispreference towards elaborate subordinators such as *in order that*, *in order to*, *so as to* (Rudnicka 2019; forthcoming)[7].

---

[6] The labels stem from the terminology of Hickey (2012: 42), according to whom, changes which are triggered and guided by social factors can be seen as *externally*-motivated changes. With regard to changes which can be tracked down to structural properties and considerations in language "and which are independent of sociolinguistic factors" Hickey (2012: 42) suggests a label of *internally*-motivated changes.

[7] Among other recent developments visible in the constructional layer and likely caused by informalization and colloquialization are e.g. the spread of the progressive tense in Present Day English (Rohe 2019); the emancipation of some semi-modals such as *gonna*/*gotta*/*wanna* (Lorenz 2013); and

Internally-motivated higher-order processes are exemplified by the rise of the *to*-infinitive, a process described in detail by Los (2005), which is linked to other trends such as i) the so-called diffusional spreading of infinitival clauses with notional subjects introduced by *for* (De Smet 2013), and ii) a trend in which non-finite clauses replace finite clauses (Mair & Leech 2006: 336). In the constructional network in question, it may be, at least to an extent, accounted for the drastic decrease in the frequency of *in order that* (see Rudnicka forthcoming) and *lest*.

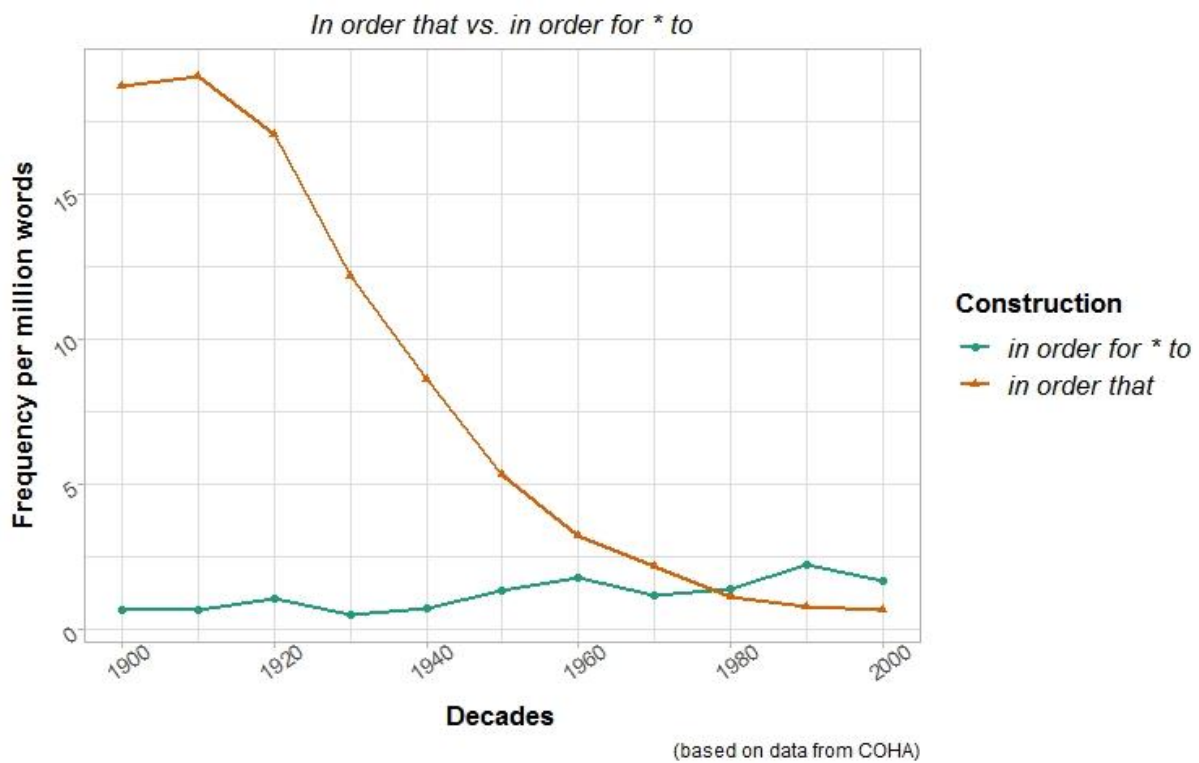

Fig. 3: Diachronic trends for *in order that* and *order for \* to* in the time period 1900-2009. (Taken from Rudnicka 2019: 95)

Along similar lines Kempf (forthcoming) relates loss to socio-cultural changes of the time such as democratization, literalization and the Enlightenment. With regard to different possible causes of the demise of *so*-relatives in New High German, she concludes (Kempf, forthcoming: 32) that "[n]either their stylistic affiliations nor their grammatical features were fatal by themselves; instead, the combination of these properties got into conflict with general grammatical, conceptual, and typological changes of the time."

---

the increase of informal linguistic variants as exemplified by the *going-to* future observed in the British parliamentary record (Hiltunen, Räikkönen, Tyrkkö 2020).

This account of possible causes is very fitting to the scenario observed by Rudnicka and exemplifies different higher-order processes at work, which Kempf acknowledges by stating "[t]his constellation strikingly parallels the scenario Rudnicka (2019: 188) develops for the loss of purpose subordinators in Late Modern English, even though the culture and language-specific developments, of course, differ."

The work by Elsweiler and Huber (forthcoming) deals with the loss of number contrast in the standard English second person pronoun. As they write, traditionally, the loss of the T pronoun was liked to stigmatization as it was associated with speech of people from rural areas, radical religious communities and lower classes. Thus it is usually categorized as a sociopragmatic change from above. Elsweiler and Huber's research (forthcoming: 21), however, points to the possibility that an additional driving force might have been a change from below, namely "the avoidance of the inflectional ending {-st} that comes along with T. Getting rid of {-st} results in a much simplified verbal paradigm". They cautiously suggest that deflexion (a general loss of inflectional morphological categories, e.g. Norde 2001) might have been one of the reasons behind the loss of T pronouns. This possibility provides another proof for looking up the constructional and supra-constructional ladder, as deflexion is an example of higher-order processes (Hilpert 2013: 14).

## 6. Summary and conclusions

Let us now come back to the statement by Haspelmath (2004: 33) quoted in Section 3, that is "[e]verything in language can become obsolete, independently of its degree of grammaticalization". The present work does not argue that this statement is not true, on the contrary, it is believed that it is true, but it does need to be supplemented by at least two pieces of information:

i) there will always be certain pre-warnings such as the presence of negative correlation between time and the frequency of use (Rudnicka 2019); distributional fragmentation (Leech et al. 2009); paradigmatic atrophy (Leech et al. 2009);

ii) only constructions which are or turn out to be not in line with one or more higher-order processes operating in a language at given time will become obsolescent or lost (Rudnicka 2019: 2011; forthcoming). In case of replacement, the replacing form will be more "in line" with the higher-order processes than the declining one.